\documentclass[10pt, letterpaper]{ieeeconf}
\usepackage[utf8]{inputenc}

\usepackage{algorithm}
\usepackage{algorithmicx}
\usepackage{algpseudocode}
\usepackage{amsmath}
\usepackage{amsfonts}
\usepackage{amssymb}
\usepackage{dsfont}

\usepackage{graphicx}
\usepackage{bbm}
\usepackage{subcaption}

\newcommand{\ep}{\hfill $\blacksquare$}

\newcommand{\E}[1]{\mathbb{E}\left[#1\right]}

\newtheorem{theorem}{Theorem}
\newtheorem{lemma}{Lemma}

\title{{Predictive Bandits}}
\author{Simon Lindståhl, Alexandre Proutiere and Andreas Johnsson
\thanks{This work was supported by the Wallenberg AI, Autonomous Systems and Software Program (WASP) funded by the Knut and Alice Wallenberg Foundation.}
\thanks{S. Lindståhl and A. Proutiere are with the Division of Decision and Control Systems, School of Electrical Engineering and Computer Science, Royal institute of Technology (KTH), Stockholm, Sweden. A. Johnsson is with Ericsson Research, Stockholm, Sweden. Emails: \{{\it lindstah@kth.se, alepro@kth.se, andreas.a.johnsson@ericsson.com}\}.}
}

\begin{document}
	\maketitle
	
\begin{abstract}
We introduce and study a new class of stochastic bandit problems, referred to as {\it predictive bandits}. In each round, the decision maker first decides whether to gather information about the rewards of particular arms (so that their rewards in this round can be predicted). These measurements are costly, and may be corrupted by noise. The decision maker then selects an arm to be actually played in the round. Predictive bandits find applications in many areas; e.g. they can be applied to channel selection problems in radio communication systems. In this paper, we provide the first theoretical results about predictive bandits, and focus on scenarios where the decision maker is allowed to measure at most one arm per round. We derive asymptotic instance-specific regret lower bounds for these problems, and develop  algorithms whose regret match these fundamental limits. We illustrate the performance of our algorithms through numerical experiments. In particular, we highlight the gains that can be achieved by using reward predictions, and investigate the impact of the noise in the corresponding measurements. 
\end{abstract}
	
	\section{Introduction}
	\label{sec:introduction}
	In this paper, we introduce and study a new class of stochastic bandit problems, referred to as {\it predictive bandits}. In the classical stochastic Multi-Armed Bandit (MAB) problem  \cite{lai1985}, the decision maker selects an arm in each round, and observes a realization of its random reward. The average rewards of the arms are initially unknown, and the objective of the decision maker is to devise a learning algorithm maximizing its reward accumulated over time. In predictive bandits, in each round, the decision maker may, before actually playing an arm, gather information about the rewards of particular arms in this round. By measuring an arm, she can {\it predict} to some extent its outcome. Measurements however come with a (fixed and known) cost, and may be corrupted by noise. As in classical stochastic MAB problems, the average rewards of the various arms are initially unknown, which forces the decision maker to explore sub-optimal arms. With predictive bandits, she has the additional difficulty of learning whether measuring arms yield better accumulated rewards, and in that case, which arms should be measured.  

Predictive bandits bear similarities with contextual bandits \cite{li2010}, where the decision maker observe {\it feature} vectors associated with each arm before playing an arm. Contextual bandits were motivated by the design of personalized recommender systems (the context may include information about both the items to be recommended and the user currently requesting a recommendation), and have been applied to the design of various web-based services. Contextual bandits differ from predictive bandits since in the latter, the observation of the context is not free, and the decision maker needs to decide which part of the context (which arm) she wishes to observe if any. Beyond web-based services and recommender systems, predictive bandits can be also applied to numerous resource allocation problems in communication networks. For example, in  the channel selection problems in radio communication systems (see e.g. \cite{chaporkar, lairadio} and references therein), the transmitter needs to choose from several radio channels, with randomly varying conditions and unknown means. One may measure the state of a channel (using probe packets) before choosing a transmission channel, but acquiring this information is consuming time and power (i.e. it has a cost). 

In this paper, we provide the first theoretical results on predictive bandits. We consider problems where the decision maker is allowed to measure at most one arm per round. In the aforementioned radio channel selection problem, such a scenario is motivated by the fact the transmitter may not have time to measure several channels without breaking the required delay guarantees of the underlying application. For predictive bandits with at most one measurement per round, our contributions are as follows.\\
(a) We derive asymptotic instance-specific regret lower bounds. These bounds constitute fundamental performance limits that no learning algorithm can beat, but they also provide insights into the design of efficient algorithms. Indeed, the lower bounds specify the optimal exploration process, i.e., the rates at which an optimal algorithm should explore sub-optimal actions. These rates depend on the average rewards of the arms and on the measurement cost. Hence, an algorithm following these exploration rates would truly and optimally adapt to the actual problem parameters.\\
(b) We present simple algorithms that rapidly learn the optimal action, and that in fact, match our regret lower bounds. These algorithms leverage KL-UCB indices \cite{lai1987adaptive} to explore sub-optimal actions, and critically rely on an aggressive exploitation strategy (in each round, the best empirical action is played with a strictly positive probability). Our main technical contribution is to establish that such an aggressive exploitation behavior is indeed asymptotically optimal. We believe that this result is general, and could be extended to many bandit problems.\\
(c) We illustrate the performance of our algorithms through numerical experiments. In particular, we highlight the gains that can be achieved by using reward predictions, and investigate the impact of the noise in the corresponding measurements.

	\section{Related Work}
	\label{sec:related}
	Stochastic bandit problems have been extensively studied. In their seminal paper \cite{lai1985}, Lai and Robbins derived asymptotic regret lower bounds and proposed algorithms achieving these fundamental limits. In \cite{auer2002}, the authors proposed UCB, a very simple and popular algorithm approaching regret lower bounds and for which a finite-time regret analysis is possible. Another attractive algorithm, KL-UCB, inherits the simplicity of UCB, and has first been shown to be asymptotically optimal in \cite{lai1987adaptive}. Later, \cite{garivier2011} proposed a finite-time analysis of the algorithm, and derived many interesting properties. 

The aforementioned papers deal with standard bandit problems, where the reward of an arm is observed only if it is played. Other types of feedback to the decision maker have been considered in the literature. In expert problems \cite{cesa2006prediction}, the rewards of all arms are observed in each round. Hybrid feedback, between the standard bandit and the expert feedback, has been analyzed in \cite{seldin14}. None of these work addresses the problem considered in this paper, where the reward of an arm (or a noisy version of it) can be observed, before actually playing an arm. 

In a recent work \cite{zuo2019observe}, the authors study a bandit problem with Bernoulli rewards where in each round, the decision maker proposes an ordered list of the $K$ arms, and plays the first arm with observed reward equal to 1. This problem is similar to that investigated in \cite{combes2015learning}. The authors devise in this setting an algorithm with regret scaling as $K^2\log(T)$. However, for this problem, it is easy to show that a constant regret (not scaling with $T$) is achievable. The problem differs from ours, since we assume that the decision maker may observe a single arm only before playing one. In addition, we consider the case of noisy measurements, and we do not restrict our attention to algorithms forced to select an arm, should its measurement returns 1 (this can be sub-optimal in the case of noisy measurements). 

Finally, it is worth mentioning contextual bandit problems \cite{li2010}, where arm {\it features} are observed as a side information to help the arm selection process. One may think that our problem falls into the class of contextual bandits -- features could be the actual arm rewards. However, here, we consider scenarios where the decision maker actively selects parts of context to be observed. Such a scenario in contextual bandits is considered in \cite{bouneffouf17}, but without any theoretical analysis.

	\section{Models and Preliminaries}
	\label{sec:problem}
	
We consider the classical stochastic bandit problem, with a set $[K]=\{1,\ldots,K\}$ of arms. The reward generated by arm $k$ in round $t\ge 1$ is denoted by $X_k(t)$. We assume that $(X_k(t))_{t\ge 1}$ is a sequence of i.i.d. random variables with Bernoulli distribution of mean $\theta_k$. Rewards are independent across arms. We denote $\theta=(\theta_1,\theta_2,...,\theta_K)$, and assume w.l.o.g. that $\theta_1>\theta_2>...>\theta_K$.  

\medskip\noindent
{\bf Measurements.} At the beginning of each round, before playing an arm, the decision maker may decide to {\it measure} an arm $k$ at a known cost $c>0$. When she decides to measure arm $k$ in round $t$, she observes the realization $Z_k(t)$ of a binary random variable, correlated with $X_k(t)$. More precisely, the observation is assumed to correspond to the output of a noisy binary channel with input $X_k(t)$, and the distribution of $Z_k(t)$ given $X_k(t)$ is: almost surely,
\begin{equation}\label{eq:channel}
\mathbb{P}[Z_k(t)=X_k(t) | X_k(t)] = 1-\varepsilon.
\end{equation}
The noise level $\varepsilon$ defines the accuracy of the measurement, and is known to the decision maker. In this paper, we consider two scenarios depending on the measurement accuracy:
\begin{itemize}
\item[(i)] {\it Perfect measurements:} $\varepsilon=0$;
\item[(ii)] {\it Imperfect measurements:} $\varepsilon \in (0,1/2)$.
\end{itemize}

\medskip
\noindent
{\bf Static policies.} A static policy (also called {\it action} in the introduction) $u$ represents the sequence of decisions made in a single round. We distinguish two types of policies. (i) Those directly playing an arm: we denote by $u=(k)$ the policy consisting in playing arm $k$. (ii) Those measuring an arm before actually playing one: such a policy $u$ is described by a triplet $(k,\ell,m)$, where $k$ is the measured arm, and $\ell$ (resp. $m$) is the arm played if the outcome of the measurement is 1 (resp. 0). We denote by ${\cal U}$ the set of static policies, and by $\mu(u)$ the average reward of policy $u$. For simplicity, we also use the notation $(k,\ell)$ to denote the policy $(k,k,\ell)$. The objective is to design an algorithm learning the optimal static policy. It is straightforward to check that the optimal policy is either $(1)$ (play the best arm without measuring) or $(1,2)$ defined as the policy consisting in measuring arm 1, in playing arm 1 if the outcome of the measurement is 1, and in playing 2 if this outcome is 0. One may also easily check that $(1,2)$ and $(2,1)$ have the same average reward (i.e., $\mu(1,2)=\mu(2,1)$). We have:
$$
\left\{
\begin{array}{l}
\mu(1) = \theta_1,\\
\mu(1,2) = -c + (1-\varepsilon)(\theta_1 + \theta_2)+ (2\varepsilon -1)\theta_1\theta_2.
\end{array}
\right.
$$
Throughout the paper, we assume that $\mu(1)\neq \mu(1,2)$. Hence the optimal static policy, denoted by $u^\star(\theta)$, is unique (when $(1,2)$ is optimal, the only other optimal policy is $(2,1)$). For simplicity, we denote $\mu^\star = \mu(u^\star(\theta))$.

\medskip
\noindent
{\bf Online learning algorithms and their regret.} An online learning algorithm $\pi$ starts with no knowledge of $\theta$, and aims at gathering data in an active manner to learn $u^\star(\theta)$ as quickly as possible. Formally, we represent the observations gathered under $\pi$ up to the beginning of round $t$ by the $\sigma$-algebra ${\cal F}_t^\pi$. In round $t$, $\pi$ selects a policy $u_t^\pi$ to be applied in round $t$; $u_t^\pi$ is a ${\cal F}_t^\pi$-measurable random variable. The set of all possible online learning algorithms is denoted $\Pi$. The performance of an algorithm $\pi\in \Pi$ is captured through its regret defined, up to round $T$, as
$$
R^\pi_\theta(T) = T \mu^\star -\sum_{t=1}^T \mathbb{E}[\mu(u_t^\pi)].
$$
The regret compares the cumulative reward collected under the learning algorithm $\pi$ to that one would collect applying the best static policy in each round; it hence quantifies the price to pay to learn $u^\star(\theta)$. We aim at devising an online algorithm with minimal regret.

	\section{Regret Lower Bounds}
	\label{sec:single_predictive}
	In this section, we derive regret lower bounds satisfied by any online learning algorithms. These bounds constitute an insightful performance benchmark for learning algorithms, but also provide guidelines into their design. We distinguish the perfect and imperfect measurement scenarios.

\subsection{Perfect measurements}

To derive lower bounds, we use classical change-of-measure arguments (refer to \cite{lai1985}, and to  \cite{garivier2019explore} for a general framework). These bounds will concern so-called {\it uniformly good} algorithms: $\pi\in \Pi$ is uniformly good if its regret satisfies for any $\theta$, $R^\pi_\theta(T) = o(T^\alpha)$ for all $\alpha>0$. Observe that such algorithms exist, since UCB applied to a bandit problem with set of 'arms' ${\cal U}$ would exhibit a regret scaling logarithmically with $T$. In the following, we denote by $I(\lambda,\lambda')$ the KL divergence between two Bernoulli distributions with respective means $\lambda$ and $\lambda'$. More generally, we denote by $KL(\nu_1||\nu_2)$ the KL-divergence between two distributions $\nu_1$ and $\nu_2$ (when it is well-defined). \medskip

\begin{theorem}
	The regret of any uniformly good algorithm $\pi\in \Pi$ satisfies: for all $\theta$, 
	$$
	\lim\inf_{T\to\infty} {R^\pi_\theta(T)\over \log(T)} \ge C(\theta),
	$$
	where $C(\theta)$ is the value of the following optimization problem:
	\begin{align*}
	\min_{\eta_u \geq 0\ \forall u \in \mathcal{U}} &\sum_{u\in \mathcal{U}}\eta_u(\mu^\star-\mu(u))\\
	\text{s.t.} \ \ \ \ &\sum_{u\in \mathcal{U}}\eta_u(\mathds{1}(u=(k))+\mathds{1}(u=(k,1))) \geq \frac{1}{I(\theta_k,\bar{\theta})}\\
	& \forall k \notin u^\star(\theta),
	\end{align*}
	where the parameter $\bar{\theta}\in[0,1]$ depends on $\theta$ as follows. \\
	\underline{Case 1:} when $u^\star(\theta)=(1,2)$ (i.e., when $c < \theta_2(1-\theta_1)$), we have $\bar{\theta}=\theta_2$. The solution of the above optimization problem is $\eta_u^\star = \sum_{k\notin u^\star(\theta)}\mathds{1}(u=(k,1))/I(\theta_k,\bar{\theta})$ and hence 
	\begin{equation*}
	C(\theta) = \sum_{k=3}^K\frac{(1-\theta_1)(\theta_2-\theta_k)}{I(\theta_k,\theta_2)}.
	\end{equation*}
	\underline{Case 2:} when $u^\star(\theta)=(1)$ (i.e., when $c > \theta_2(1-\theta_1)$), we have $\bar{\theta}=\min(\theta_1,\frac{c}{1-\theta_1})$. The solution of the optimization problem is for any $k\notin u^\star(\theta)$, and any $u$ such that $k\in u$, 
	$$
	\eta_u^\star={1\over I(\theta_k,\bar{\theta})}\times \left\{
	\begin{array}{ll}
	\mathds{1}(u=(k)) & \hbox{if } c < \theta_1(1-\theta_k),\\
	\mathds{1}(u=(k,1)) & \hbox{otherwise}.
	\end{array}
	\right.
	$$
	Hence $C(\theta) = \sum_{k=2}^{K} H_k(\theta)$ with		
	\begin{equation*}
	H_k(\theta)  = {1\over I(\theta_k,\bar{\theta})}\times \begin{cases}
	c-(1-\theta_1)\theta_k, & \hbox{if }c< \theta_1(1-\theta_k) \\
	\theta_1 - \theta_k, & \text{otherwise}.
	\end{cases}
	\end{equation*}
	\label{the:CMOsolution}
\end{theorem}

In the above theorem, the solution $\eta^\star$ to the optimization problem leading to $C(\theta)$ may be interpreted as follows: $\eta_u^\star \log(T)$ is the expected number of rounds the policy $u$ should be selected by a learning algorithm minimizing regret. Such an optimal algorithm would explore only very specific policies. Indeed, for any $k\notin u^\star(\theta)$, one and only one of the policies $u=(k)$ or $u=(k,1)$ should be explored a number of rounds of the order $\log(T)$; all other policies have to be explored $o(\log(T))$ times. 

Also observe that $\bar{\theta}$ may be interpreted as the value to which the parameter $\theta_k$ should be changed to make a policy using arm $k$ (i.e. $(k)$ or $(k,1)$) optimal. Now let $\nu_\theta(u)$ denote the distribution of the observation made in a given round under the policy $u$. As this will be come clear in the proof of the theorem, the quantity $I(\theta_k,\bar{\theta})$ is actually equal to $KL(\nu_\theta(u)|| \nu_{\theta'}(u))$ for $u=(k)$ or $u=(k,1)$, where $\theta'$ is such that $\theta'_j=\theta_j$, for all $j\neq k$, and $\theta_k'=\bar{\theta}$. It can be interpreted as the {\it amount of information} brought by policy $u$ in a single round to decide whether $k$ is part of the optimal policy. It can be verified that the policy $u$ including $k$ that should be explored is the one minimizing the ratio of its regret $\mu^\star-\mu(u)$ to the amount of information brought to decide whether $k$ is part of the optimal policy. This principle is general, and will also hold in the case of imperfect measurements.

\medskip
\noindent

{\it Proof of Theorem \ref{the:CMOsolution}.} We use change-of-measure arguments. Let $\pi$ be a uniformly good algorithm. Denote by $\Lambda(\theta)$ the set of confusing problem parameters, i.e., those leading to a different optimal policy, and that cannot be distinguished from the true parameters if the optimal policy is always played. In other words:
\begin{equation}
\Lambda(\theta)=\{\lambda: KL(\nu_\theta(u^\star(\theta))|| \nu_\lambda(u^\star(\theta)))=0, u^\star(\theta)\neq u^\star(\lambda)\}.
\label{eq:LambdaDef}
\end{equation}
Note that if $u^\star(\theta))=(1)$, then 
$$
KL(\nu_\theta(u^\star(\theta))|| \nu_\lambda(u^\star(\theta)))=0 \Longleftrightarrow \lambda_1=\theta_1,
$$
and if $u^\star(\theta))=(1,2)$, then 
$$
KL(\nu_\theta(u^\star(\theta))|| \nu_\lambda(u^\star(\theta)))=0 \Longleftrightarrow (\lambda_1=\theta_1, \lambda_2=\theta_2).
$$
If $\mathbb{E}^\pi[N_u(T)]$ is the expected number of rounds where $\pi$ applies policy $u$ up to time $T$, we can show as in  \cite{garivier2019explore} that: for all $\lambda\in \Lambda(\theta)$, 
\begin{equation}
\sum_u \mathbb{E}^\pi[N_u(T)] KL(\nu_\theta(u)||\nu_\lambda(u)) \ge \log(T)(1+o(1)).
\end{equation}
Since $R^\pi_\theta(T)=\sum_u \mathbb{E}^\pi[N_u(T)](\mu^\star-\mu(u))$, this implies that an asymptotic lower bound for the regret is $C(\theta)\log(T)$, where $C(\theta)$ is the value of the solution of the following optimization problem.
\begin{align}
\min_{\eta_u \geq 0\ \forall u \in \mathcal{U}} &\sum_{u\in \mathcal{U}}\eta_u(\mu^\star-\mu(u))\label{eq:raw1}\\
\text{s.t.} \ \ \ \ &\sum_{u\in \mathcal{U}}\eta_u KL(\nu_\theta(u)||\nu_\lambda(u))\ge 1, \forall \lambda\in \Lambda(\theta)\label{eq:raw2}.
\end{align}
{\it Step 1. Pruning constraints.} We argue that we can restrict the set of constraints in the above problem, by restricting the attention to $\lambda\in \Lambda(\theta)$ such that only one coordinate of $\lambda$ differs from those of $\theta$. We distinguish two cases. First, if $(1,2)$ is the optimal policy under $\theta$, then we have $\lambda_1=\theta_1$ and $\lambda_2=\theta_2$. If under $\lambda$, $(k)$ is optimal (for $k\ge 3$), then it is easy to see that $\lambda_k$ should be set just above $\theta_2$, and we do not need to change any other component of $\theta$. Similarly, if under $\lambda$, $(k,1)$ is optimal, then changing only $\lambda_k$ is required. Now assume that under $\lambda$, $(k,\ell)$ is optimal for $k,\ell\notin \{1,2\}$. We must have: $\lambda_k+\lambda_\ell -\lambda_k\lambda_\ell > \theta_1+\theta_2-\theta_1\theta_2$, from which we deduce that either $\lambda_k$ or $\lambda_\ell$ is greater than $\theta_2$. Hence, the constraint generated by this $\lambda$ is not active. We can do the same reasoning to show that if $(1)$ is optimal under $\theta$, then the active constraints are those corresponding to $\lambda$'s that differ from $\theta$ by one coordinate only. In this case, however, it suffices that $\lambda_k > \frac{c}{1-\theta_1}$, as this will imply $\lambda_1+(1-\lambda_1)\lambda_k-c>\theta_1$. 

\medskip
\noindent
\underline{\it Step 2. Solution of (\ref{eq:raw1})-(\ref{eq:raw2}).} By studying the average rewards $\mu(u)$ and the KL-divergence $KL(\nu_\theta(u)|| \nu_\lambda(u))$ of the various policies, we can show that the solution $\eta^\star$ of (\ref{eq:raw1})-(\ref{eq:raw2}) is such that for most policies $u$, $\eta_u^\star=0$. We do so by showing that for such $u$ and for any feasible solution $\eta$, $\eta_u>0 \implies \eta \neq \eta^\star$. Assume first that $u^\star(\theta)=(1,2)$. Let $k\ge 3$. Then the set of constraints for $\lambda\in \Lambda(\theta)$ such that $\lambda_\ell=\theta_\ell$ for all $\ell\neq k$ reduces to the single constraint
$$
\sum_{u: k\in u}\eta_u KL(\nu_\theta(u)||\nu_\lambda(u))\ge 1,
$$
where $\lambda_k=\theta_2$. The KL divergences involved in this constraint are: for $\ell,\ell_1,\ell_2\neq k$,
$$
KL(\nu_\theta(u)|| \nu_\lambda(u))=\left\{
\begin{array}{ll}
I(\theta_k,\theta_2) & \hbox{ case I,} \\
(1-\theta_\ell)I(\theta_k,\theta_2) & \hbox{ case II,}\\
\theta_{\ell_1}I(\theta_k,\theta_2) & \hbox{ case III,} 
\end{array}
\right.
$$
where case I holds for $u=(k), (k,\ell), (k,\ell_1,\ell_2)$, case II for $u=(\ell,k)$, and case III for $u=(\ell_1,k,\ell_2)$. Consider $u=(k,\ell)$ with $\ell \neq k$ and $\ell > 1$, take any feasible solution $\eta$ such that $\eta_u > 0$, and consider another feasible solution $\eta'$, identical to $\eta$ except $\eta_u=0$, $\eta_{(k,1)}'=\eta_{(k,1)}+\eta_u$ and $\eta_{(\ell,1)}'=\eta_{(\ell,1)}+(1-\theta_k)\eta_u$. Then the difference in cost function between $\eta$ and $\eta'$ is
\begin{align*}
\eta_u((\mu^\star & -\mu((k,\ell))-(\mu^\star-\mu(k,1))-(1-\theta_k)(\mu^\star-\mu(\ell,1)))\\
&= \eta_u(1-\theta_k)((\theta_1-\theta_\ell)-(\theta_2-\theta_\ell)) > 0.
\end{align*}
Therefore $\eta^\star \neq \eta$ and hence $\eta_{(k,\ell)}^\star=0$.  Similar arguments lead to $\eta^\star_{(\ell,k)}=\eta_{(k,\ell_1,\ell_2)}^\star=\eta_{(\ell_1,k,\ell_2)}^\star=0$ for any $\ell, \ell_1, \ell_2 \neq k$, and, comparing $(k)$ to $(1,k)$, $\eta_{(k)}^\star=0$. By process of elimination, we deduce the results of Case 1 in Theorem \ref{the:CMOsolution}. Assuming now that $u^\star(\theta)=(1)$, we prove the results of Case 2 in Theorem \ref{the:CMOsolution} using the same arguments.
\ep

\subsection{Imperfect measurements}

%

%
%

The following theorem provides regret lower bounds in the case of imperfect measurements. For simplicity, we define $p_0(\theta_k):=\mathbb{P}[Z_k(t)=0]=\varepsilon\theta_k + (1-\varepsilon)(1-\theta_k)$.  \medskip

\begin{theorem}
	The regret of any uniformly good algorithm $\pi\in \Pi$ satisfies: for all $\theta$, 
	$$
	\lim\inf_{T\to\infty} {R^\pi_\theta(T)\over \log(T)} \ge C_\varepsilon(\theta),
	$$
	where $C_\varepsilon(\theta)$ is the value of the following optimization problem:
	\begin{align*}
	&\min_{\eta_u \geq 0\ \forall u \in \mathcal{U}} \sum_{u\in \mathcal{U}}\eta_u(\mu^\star-\mu(u))\\
	&\text{s.t. } \sum_{u\in \mathcal{U}_1(k)}\eta_u KL(\nu_{\theta}(u)||\nu_{(\theta^{(-k)},\bar{\theta})}(u)) \geq 1, \ \forall k \notin u^\star(\theta),
	\end{align*}
	where $\mathcal{U}_1(k)=\{(k),(k,1),(k,1,k), (k,1,1)\}$, and where $(\theta^{(-k)},\bar{\theta})=\theta'$ corresponds to arm rewards such that $\theta_j'=\theta_j$, for $j\neq k$, and $\theta_k'=\bar{\theta}$. The parameter $\bar{\theta}$ depends on $\theta$ as follows. When $u^*(\theta)=(1,2)$, we have $\bar{\theta}=\theta_{2}$; when $u^*(\theta)=(1)$, $\bar{\theta}=\min(\theta_1,\frac{c+\varepsilon\theta_1}{p_0(\theta_1)})$.
	
	The solution $\eta^\star$ of the above optimization problem is:
	\begin{equation}
		\eta_u^\star = \sum_{k\notin u^\star(\theta)} {
			\mathds{1}(u=u^\star_k)\over KL(\nu_{\theta}(u)||\nu_{(\theta^{(-k)},\bar{\theta})}(u))}\label{eq:noisy_solution}
	\end{equation}

	where $u^\star_k=\arg\min_{u \in \mathcal{U}_1(k)} h_k(u)$ with
	$$
	h_k(u)= {\mu^\star-\mu(u)\over KL(\nu_{\theta}(u)||\nu_{(\theta^{(-k)},\bar{\theta})}(u))}.
	$$
	Thus, $C_\varepsilon(\theta)=\sum_{k\notin u^\star(\theta)} h_k(u^\star_k)$.

	%
	%
	\label{the:NCMO-solution}
\end{theorem}

Theorem \ref{the:NCMO-solution} and its interpretation are very similar to Theorem \ref{the:CMOsolution}, and in fact Theorem \ref{the:NCMO-solution} reduces to Theorem \ref{the:CMOsolution} when $\varepsilon \to 0$, with $H_k(\theta) = h_k(u_k^\star)$. In particular, we still have that policies including more than one suboptimal arm will not be considered for exploration. To decide whether $k$ belongs to the optimal policy, an optimal algorithm should explore a single policy containing arm $k$ and possibly arm 1. However, in the case of imperfect measurements, this policy can be any of the 4 policies in ${\cal U}_1(k)$, depending on the parameter $\theta$. Again this policy is the one minimizing the ratio $h_k(u)$ of its regret to the amount of information it brings. The full proof is similar to that of Theorem \ref{the:CMOsolution} and can be found in Appendix \ref{appx:single_predictive_lower_bound}.

\section{Algorithms}	
	
In this section, we exploit our regret lower bounds to devise algorithms, in both scenarios, with perfect and imperfect measurements. We also provide an analysis of the regret of the proposed algorithms.

\subsection{Perfect measurements}

We present Single Predictive Arm Measurements (SPAM), an algorithm whose regret matches the lower bound derived in Theorem \ref{the:CMOsolution}. SPAM maintains a {\it leading} arm $j_1(t)$ defined as the best empirical arm up to round $t$, $j_1(t) \in \arg\max_k \hat{\theta}_k(t)$ (ties are broken arbitrarily), where $\hat{\theta}_k(t)$ denotes the empirical reward of arm $k$ averaged over the $(t-1)$ first rounds. It also maintains $j_2(t)$, the second best empirical arm, as well as the best empirical policy ${\cal L}(t)$ (either $(j_1(t))$ or $(j_1(t),j_2(t))$). SPAM uses KL-UCB indices: for arm $k$, this index is defined as: 
\begin{equation}
b_k(t) := \max \{q: n_k(t)I(\hat{\theta}_k,q)\leq f(t)\},
\end{equation}
where $f(t)=\log(t)+4\log\log(t)$ and $n_k(t)$ is the number of times arm $k$ has been observed up to time $t$. In each round, to decide whether SPAM should explore apparently sub-optimal policies, these indices are compared to an estimated threshold $\hat{\bar{\theta}}(t)$, equal to $\hat{\theta}_{j_2(t)}(t)$ if $\hat{\theta}_{j_2(t)}(t)\geq \frac{c}{1-\hat{\theta}_{j_1(t)}(t)}$ and $\min(\hat{\theta}_{j_1(t)}(t),\frac{c}{1-\hat{\theta}_{j_1(t)}(t)})$ otherwise. SPAM only explores policies containing arms in the following set of uncertain arms:
 \begin{equation}
\mathcal{B}(t) := \{k: b_k(t) \geq \hat{\bar{\theta}}\}.
\end{equation}
SPAM {\it exploits}, i.e., select the leading policy ${\cal L}(t)$, very regularly (with probability at least 1/2 in each round), so that the arms in the leading policy are very well estimated. SPAM {\it explores} apparently sub-optimal policies only if the set ${\cal B}(t)$ is not empty. More precisely, it explores either $(k)$ or $(k,j_1(t))$ for $k\in {\cal B}(t)$. All the design choices made in SPAM are aligned to the optimal exploration process suggested in our regret lower bound. The pseudo-code of SPAM is presented in Algorithm \ref{alg:SPAM}.
	
	\begin{algorithm}[h!]
		\caption{\textsc{SPAM}}
		\label{alg:SPAM}
		\begin{algorithmic}[1]
			\State Initialize $\hat{\theta}_k(1)=1$ and $b_k(1)=1$ for all arms $k$, \\
			${\cal B}(1)=\emptyset$, and ${\cal L}(1)$ arbitrarily.
			\For{$t = 1,2,...$}
			\If{$\mathcal{B}(t) = \emptyset$} exploit: $u\leftarrow \mathcal{L}(t)$,
			\Else
			\State w.p. $1/2$, exploit: $u\leftarrow \mathcal{L}(t)$,
			\State w.p. $1/2$, explore: choose $k$ uniformly at random
			\State \ \ \ \ from $\mathcal{B}(t)$, then:
			\State \ \ \ \ \ \ \ $u\leftarrow (k,j_1(t))$ if $(1-\hat{\theta}_{j_1(t)}(t))\hat{\theta}_k(t)>c$,
			\State \ \ \ \ \ \ \ $u\leftarrow (k)$ otherwise.
			\EndIf
			\State Play policy $u$ and observe its outcomes.
			\State Compute $\hat{\theta}_k(t+1)$ and $b_k(t+1)$ for all arms $k$,
			\State Compute $\mathcal{B}(t+1)$, $\mathcal{L}(t+1)$.
			
			\EndFor
		\end{algorithmic}
		
	\end{algorithm}
	
Before we provide, in the theorem below, a finite-time analysis of the regret of SPAM, we introduce the following notations. For any $\theta$, let $\delta_0$ be such that (i) $\delta_0 \leq \min_{i<K}(\frac{1}{2}(\theta_i-\theta_{i+1}))$, (ii) if $u^\star(\theta)=(1,2)$, $\frac{c}{1-\theta_1-\delta_0}\leq\theta_2-\delta_0$, and (iii)
if $u^\star=(1)$, $\frac{c}{1-\theta_1+\delta_0}\geq \theta_2 + \delta_0$. It can be easily checked that such a $\delta_0$ indeed exists. Let $\beta=(1-\theta_1)^{-1}$, and define for $\delta>0$,
\begin{equation*}
g(\theta_1,\theta_2,\delta) := \begin{cases}
\theta_1 - \delta, &\bar{\theta}=\theta_{1}\\
\theta_2 - \delta, &\bar{\theta}=\theta_{2}\\
\frac{c}{1-\theta_{1}+\delta}, &\bar{\theta}=\frac{c}{1-\theta_1}.
\end{cases}
\end{equation*}
Finally, we introduce the functions $H_k$ so that the constant $C(\theta)$ involved in regret lower bound derived in Theorem \ref{the:CMOsolution} can be written as $C(\theta)=\sum_{k\notin u^\star(\theta)}H_k(\theta)$ in all cases. Hence, if $u^\star=(1,2)$, we have $H_k(\theta):= (1-\theta_1)(\theta_2=\theta_k)/I(\theta_k,\theta_2)$ and if $u^\star=(1)$, $H_k(\theta)$ is defined as in Theorem \ref{the:CMOsolution}. \medskip
	
\begin{theorem}
There is a constant $c>0$ such that for any $\theta$, any $\delta < \delta_0$ and any $\epsilon \in (0,1/2)$, the regret of SPAM satisfies: for all $T\ge 1$,
\begin{align}
R^{\mathrm{SPAM}}_\theta(T) \le & \sum_{k\notin u^\star(\theta)}{H_k(\theta)I(\theta_k,\bar{\theta})\over (1-\epsilon)I(\theta_k,g(\theta_1,\theta_2,\delta))} f(T)\nonumber\\
&\ \ \ \ \  +c K(K+\beta^2) + \epsilon^{-2} +\delta^{-2}(\beta+1).
\end{align}
\label{the:SinglePredictiveUpperBound}
\end{theorem}
	
An immediate consequence of the above theorem, whose proof can be found in Appendix \ref{appx:single_predictive_upper_bound}, is that SPAM is asymptotically optimal. Indeed, by letting first $T$ tend to $\infty$, and then $\epsilon, \delta$ to 0, we obtain:
$$
\limsup_{T\to\infty} {R^{\mathrm{SPAM}}_\theta(T)\over \log(T)} \le C(\theta).
$$
 	
\subsection{Imperfect measurements}

The design of our algorithm for the case of noisy measurements follows the same principles as that of SPAM, but is slightly complicated because: (i) According to our lower bounds, to determine whether arm $k$ belongs to the optimal policy, the 4 policies of ${\cal U}_1(k)$ could be used in the exploration process. (ii) Due to the noisy measurements, the estimation of $\theta_k$ is slightly involved. Next, we propose Noisy Single Predictive Arm Measurements (NoSPAM), an extension of SPAM to the case of noisy measurements. The regret analysis of NoSPAM is complicated by the aforementioned facts. We believe that NoSPAM is asymptotically optimal, just as SPAM, but omit the analysis here. The main difference between SPAM and NoSPAM lies in the estimation of the parameters $\theta$, which we explain next.

\begin{figure*}[h]
	\centering
	\begin{subfigure}{0.3\textwidth}
		\centering
		\includegraphics[width=\textwidth]{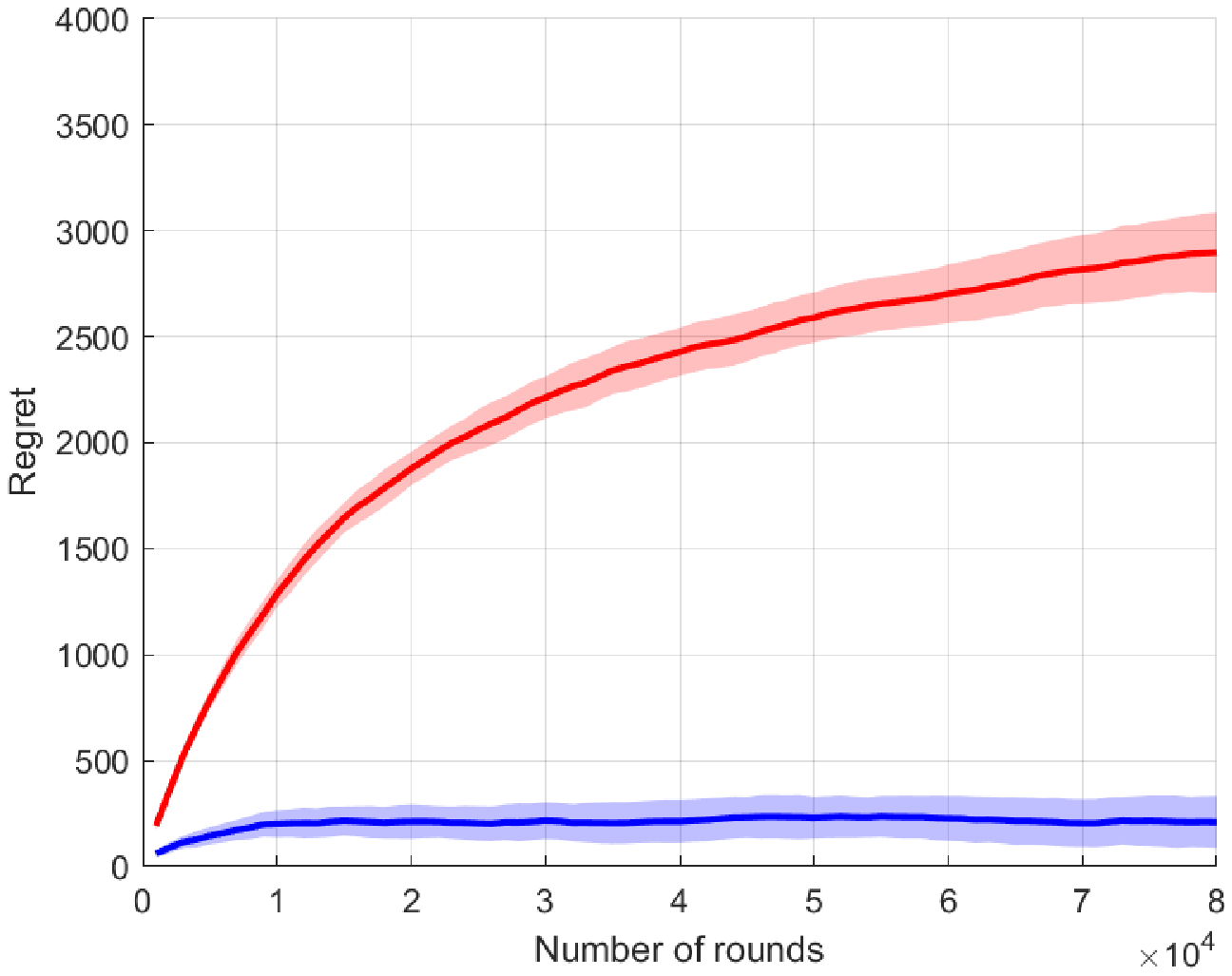}
		\caption{\textsc{SPAM}}
		\label{fig:NoiselessResults}
	\end{subfigure}
	\begin{subfigure}{0.3\textwidth}
		\centering
		\includegraphics[width=\textwidth]{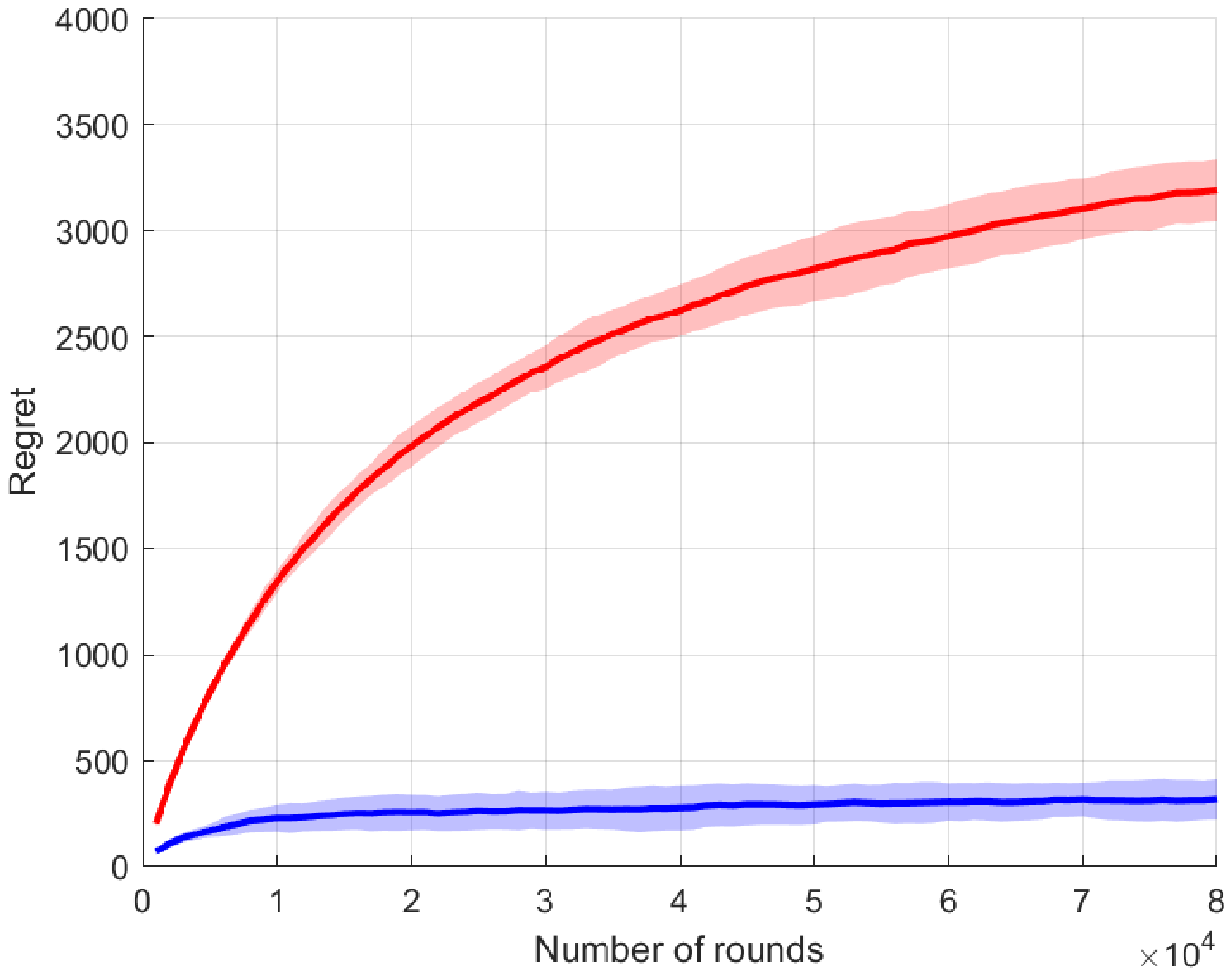}
		\caption{\textsc{NoSPAM} with $\varepsilon=0.1$}
		\label{fig:NoisyResults}
	\end{subfigure}
	\begin{subfigure}{0.3\textwidth}
		\centering
		\includegraphics[width=\textwidth]{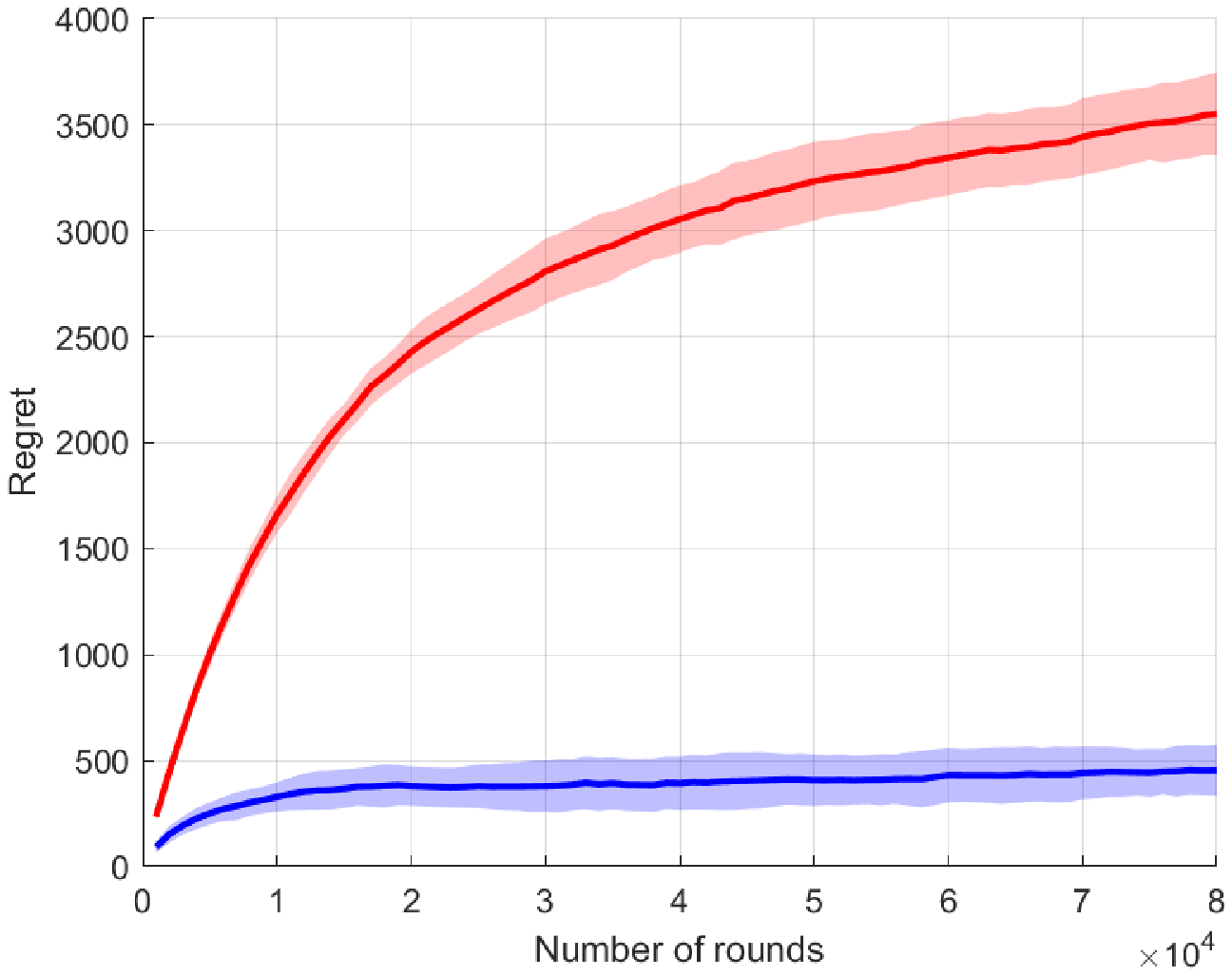}
		\caption{\textsc{NoSPAM} with $\varepsilon=0.3$}
		\label{fig:VeryNoisyResults}
	\end{subfigure}
	\caption{Regret of our algorithms (blue) compared to that of an unstructured \textsc{KL-UCB} algorithm (red) for different values of $\varepsilon$. The shaded areas correspond to one standard deviation.}
	\label{fig:Results}
	
\end{figure*}

\medskip
\noindent
{\bf Estimating average arm rewards.} To derive $\hat{\theta}_k(t)$, an estimator of $\theta_k$, we use the following quantities. Let $n_{1,k}(t)$ be the number of rounds $s$ up to round $t$ where it has been observed that $X_k(s)=1$; let $n_{2,k}(t)$ be the number of rounds $s$ where it has been $X_k(s)=0$; let $n_{3,k}(t)$ be the number of rounds $s$ where $Z_k(s)=1$ has been observed but $X_k(s)$ has not been observed, and finally let $n_{4,k}(t)$ be the number of rounds $s$ where $Z_k(s)=0$ has been observed but $X_k(s)$ has not been observed. Define $n_k(t)=\sum_{i=1}^{4} N_{i,k}(t)$,  the number of rounds $s$ where either $Z_k(s)$ or $X_k(s)$ have been observed observed. It can be readily shown that the maximum-likelihood estimator $\hat{\theta}_k$ of $\theta_k$ is the solution $X\in [0,1]$ of the following cubic equation: $a_1 X^3+a_2 X^2 +a_3X+a_4 =0$, where
$$
\left\{
\begin{array}{l}
a_1= n_k(t)(1-2\varepsilon)^2,\\
a_2= (1-2\varepsilon)\big(\varepsilon n_{k,3}(t)-(1-\varepsilon)n_{k,4}(t)\\
\ \ \ \ -(1-2\varepsilon)(n_k(t)+n_{k,1}(t))\big),\\
a_3 = -\varepsilon(1-\varepsilon)n_k(t)+(1-2\varepsilon)^2 n_{1,k}(t)\\
\ \ \ \ +\varepsilon^2 n_{3,k}(t)+(1-\varepsilon)^2 n_{4,k}(t),\\
a_4 = \varepsilon(1-\varepsilon)n_{1,k}(t).
\end{array}
\right.
$$
	\begin{algorithm}[h!]
	\caption{\textsc{NoSPAM}}
	\label{alg:NoSPAM}
	\begin{algorithmic}[1]
		\State Initialize $\hat{\theta}_k(1)=1$ and $b_k(1)=1$ for all arms $k$, \\
		${\cal B}(1)=\emptyset$, and ${\cal L}(1)$ arbitrarily.
		\For{$t = 1,2,...$}
		\If{$\mathcal{B}(t) = \emptyset$} exploit: $u\leftarrow \mathcal{L}(t)$,
		\Else
		\State w.p. $1/2$, exploit: $u\leftarrow \mathcal{L}(t)$,
		\State w.p. $1/2$, explore: choose $k$ uniformly at random
		\State \ \ \ \ from $\mathcal{B}(t)$, then:
		\State \ \ \ \ \ \ \ for $u\in \Gamma :=\{ (k,j_1(t)), (k), (k,j_1(t),k),$\\ 
		\ \ \ \ \ \ \ \ \ \ \ \ \ \ \ \ \ \ \ \ \ \ $(k,j_1(t),j_1(t))\}$, calculate 
		\State  \ \ \ \ \ \ \ \ \ \ \ \  $h_k(u)=\frac{\mu(\mathcal{L}(t))-\mu(u)}{KL(\nu_{\hat{\theta}(t)}(u)||\nu_{(\hat{\theta}^{(-k)}(t),\hat{\bar{\theta}}(t))}(u))}$
		\State \ \ \ \ \ \ \ $u\leftarrow \arg\min_{u\in \Gamma} h_k(u)$
		\EndIf
		\State Play policy $u$ and observe its outcomes.
		\State Compute $\hat{\theta}_k(t+1)$ and $b_k(t+1)$ for all arms $k$,
		\State Compute $\mathcal{B}(t+1)$, $\mathcal{L}(t+1)$.
		
		\EndFor
	\end{algorithmic}
	
\end{algorithm}

\medskip
\noindent	
Now, defining $n_{k}^{\mathrm{play}}(t)$ (resp. $n_u(t)$) as the number of rounds $s$ up to $t$ where $X_k(s)$ is observed but not $Z_k(s)$ (resp. $u$ is selected), we can define the KL-UCB index of arm $k$ as: 

\begin{align*}
b_k(t)&= \max\{q: \sum_{u \in \mathcal{U}_m(k)}n_u(t)KL(\nu_{\hat{\theta}(t)}(u)||\nu_{(\hat{\theta}^{(-k)}(t),q)}(u))\\
&\ \ \ \ +n_{k}^{\mathrm{play}}(t)I(\hat{\theta}_k(t),q) \leq f(t)\},
\end{align*}
where $f(t)=\log(t)+4\log(\log(t))$ and $\mathcal{U}_m(k)$ denotes the set of all policies where $k$ is measured. Recall that $\nu_{(\hat{\theta}^{(-k)}(t),q)}(u)$ is defined in Theorem \ref{the:NCMO-solution}. $\mathcal{B}(t)$, $j_1(t)$, $j_2(t)$ and $\mathcal{L}(t)$ are defined as for \textsc{SPAM}, with $\hat{\bar{\theta}}(t)=\hat{\theta}_{j_2(t)}(t)$ if $\hat{\theta}_{j_2(t)}(t)\geq \frac{c+\varepsilon\hat{\theta}_{j_1(t)}(t)}{p_0(\hat{\theta}_{j_1(t)}(t))}$ and $\hat{\bar{\theta}}(t)=\min(\hat{\theta}_{j_1(t)}, \frac{c+\varepsilon\hat{\theta}_{j_1(t)}(t)}{p_0(\hat{\theta}_{j_1(t)}(t))})$ otherwise. The pseudo-code of \textsc{NoSPAM} is presented in Algorithm \ref{alg:NoSPAM} below.

	
	
	\section{Numerical Experiments}
	\label{sec:experimental}

\begin{figure}[t]
	\centering
	\includegraphics[width=0.4\textwidth]{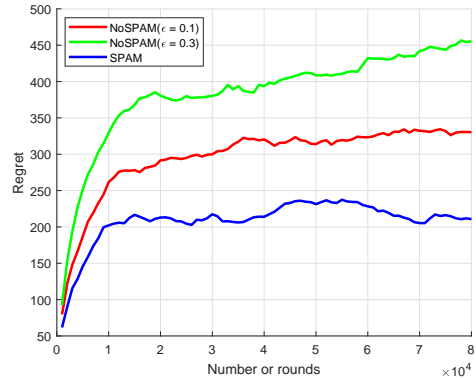}
	\caption{Regret of \textsc{SPAM} and \textsc{NoSPAM} for various $\varepsilon$.}
	\label{fig:EpsComparison}
\end{figure}

%

In this section, we illustrate the performance of SPAM and NoSPAM. We compare their performance to that of KL-UCB when applied to the set of static policies (as if this was the set of arms). KL-UCB is known to be asymptotically optimal when the various arms have uncorrelated rewards. Here, however, the rewards of policies using the same arm are correlated, and this is precisely this structure that SPAM and NoSPAM optimally exploit.  

\medskip
\noindent
{\bf Implementation of KL-UCB.} KL-UCB selects the policy with the highest KL-UCB index. We know a priori that the optimal static policy is of the form $(k)$ or $(k,\ell)$, and so naturally, we restrict KL-UCB to these policies. To exploit all the observations made up to round $t$, we define the KL-UCB index of policy $u$ as: If $u=(k)$, $b_u(t)=\max\{q:n_u(t) I(\hat{\theta}_k(t),q)\}$, and if $u=(k,\ell)$,
\begin{equation*}
b_u(t) = \max\{q: n_u(t)I((1-\varepsilon)\hat{\theta}_{k}(t)+p_0(\hat{\theta}_k(t))\hat{\theta}_\ell(t),q)\}-c.
\end{equation*}
Here $n_u(t)$ is the number of times policy $u$ has been played up to time $t$. The $\hat{\theta}_k(t)$'s are computed as in Algorithm \ref{alg:SPAM}.

\medskip
\noindent
{\bf Experiment setup.} We run an experiment with $K=10$ arms with expected rewards $\theta_k = 0.55\cdot (1-(k-1)/(K-1))$ for $k=1,\ldots,10$. the measurement cost is fixed to $c=0.1$. The time horizon is $T=8\cdot 10^4$ and we average the regret over $20$ runs. We test \textsc{SPAM} (with $\epsilon=0$) as well as \textsc{NoSPAM} with $\varepsilon=0.1$ and $\varepsilon=0.3$. The results of these experiments are reported in Figure \ref{fig:Results}. Finally, we present the regret of \textsc{SPAM} and \textsc{NoSPAM} in the same plot in Figure \ref{fig:EpsComparison} to visualize the impact of increasing the noise level on regret.

As expected, Figure \ref{fig:Results} shows that \textsc{SPAM} and NoSPAM vastly outperform the unstructured \textsc{KL-UCB} (with or without noise). Figure \ref{fig:EpsComparison} shows that there is a similar loss in regret when moving from $\varepsilon=0.1$ to $\varepsilon=0.3$ as when moving from $\varepsilon=0$ to $\varepsilon=0.1$, suggesting that \textsc{NoSPAM} is indeed a natural extension of \textsc{SPAM}.

	\section{Conclusion}
	\label{sec:conclusion}
	In existing bandit and contextual bandit problems, the decision maker cannot decide to observe the rewards of specific arms or their contexts before actually playing an arm. Such an observation in a given round would help the decision maker to predict the rewards in that round, but would typically come with a cost. In this paper, we move towards such {\it predictive} bandits and investigate problems where the agent can measure the reward of at most one arm before making playing an arm. These measurements are either perfect or have a known probability of being incorrect. We derive a regret lower bound for these problems, and devise algorithms in the endeavor of matching these bounds. This paper proposes the first analytical results on predictive bandits, and naturally suggests interesting research directions. We can for instance extend the analysis to problems where the agent may measure multiple arms. More generally, it would be also interesting to investigate contextual bandit problems where the agent must choose which parts of the context to observe.

	
	\appendices
	
	
	

	\section{Proof of Theorem \ref{the:NCMO-solution}}
	\label{appx:single_predictive_lower_bound}
	
\begin{proof}
	We use a similar argument as in the proof of Theorem \ref{the:CMOsolution}.  Recall that $\nu_{\theta}(u)$ denotes the distribution of the random observation when under policy $u$. It is easy to see that if $u=(k)$, the mapping $\theta_k\to\nu_\theta(u)$ is one to one, and if $u=(k,\ell,m)$ the mapping $(\theta_k,\theta_\ell,\theta_m)\to \nu_\theta(u)$ is one-to-one. Denote by $\Lambda(\theta)$ the set of confusing parameters, defined in equation \eqref{eq:LambdaDef}. Since we have one-to-one mappings, it is again true that if $u^*(\theta)=(1)$ we have $\lambda_1=\theta_1$ and if $u^*(\theta)=(1,2)$ we have $\lambda_1=\theta_1$, $\lambda_2=\theta_2$. We furthermore have that if $\mathbb{E}^\pi[N_u(T)]$ is the expected number of rounds where $\pi$ applies policy $u$ up to time $T$, for all $\lambda \in \Lambda(\theta)$ and for large $T$
	\begin{equation}
	\sum_u \mathbb{E}^\pi[N_u(T)] KL(\nu_\theta(u)||\nu_\lambda(u)) \ge \log(T)(1+o(1)).
	\end{equation}
	This implies that an asymptotic lower bound for the regret is $C(\theta)\log(T)$, where $C(\theta)$ is the value of the solution of the following optimization problem.
	\begin{align}\label{eq:noisyraw}
	\min_{\eta_u \geq 0\ \forall u \in \mathcal{U}} &\sum_{u\in \mathcal{U}}\eta_u(\mu^\star-\mu(u))\\
	\text{s.t.} \ \ \ \ &\sum_{u\in \mathcal{U}}\eta_u KL(\nu_\theta(u)||\nu_\lambda(u))\ge 1, \forall \lambda\in \Lambda(\theta).
	\end{align}
	We can show, by precisely the same reasoning as in the proof of Theorem \ref{the:CMOsolution}, that it is enough to consider $\lambda$ which differ from $\theta$ in only one component $\lambda_k$ (with $k \notin u^\star$) and with $\lambda_k>\bar{\theta}$. It thus remains to solve the optimization problem \eqref{eq:noisyraw}.\medskip
	
	We will show that there exists an optimal solution $\eta$ such that, for most cases, $\eta_u=0$. We will treat the case $u^\star(\theta)=(1,2)$, the case $u^*(\theta)=(1)$ will be analogous. First, take $u=(k,\ell,m)$ with $m\neq k \neq \ell \neq m$ and $k>1$, $l>1$, $m>1$. For any feasible solution $\eta$ such that $\eta_u>0$, take the related, also feasible, solution $\eta'$ identical to $\eta$ except $\eta_u'=0$, $\eta_{(k,1,1)}'=\eta_{(k,1,1)}+\eta_u$, $\eta_{(1,l)}'=\eta_{(1,l)}+\frac{1-p_0(\theta_k)}{p_0(\theta_1)}\eta_u$, $\eta_{(1,m)}'=\eta_{(1,m)}+\frac{p_0(\theta_k)}{p_0(\theta_1)}\eta_u$. Then, the difference between the cost functions of $\eta$ and $\eta'$ is $\eta_u$ multiplied by
	\begin{align*}
		&(\mu^\star-\mu(k,\ell,m))-(\mu^\star-\mu(k,1,1))\\
		&\ \ \ \  -\frac{1-p_0(\theta_k)}{p_0(\theta_1)}(\mu^\star-\mu(1,\ell))-\frac{p_0(\theta_k)}{p_0(\theta_1)}(\mu^\star-\mu(1,m))\\
		&=\mu(k,1,1)-\mu(k,\ell,m)\\
		&\ \ \ \  -\frac{1-p_0(\theta_k)}{p_0(\theta_1)}p_0(\theta_1)(\theta_2-\theta_\ell)-\frac{p_0(\theta_k)}{p_0(\theta_1)}p_0(\theta_1)(\theta_2-\theta_m)\\
		&=(1-p_0(\theta_k))(\theta_1-\theta_\ell- (\theta_2-\theta_\ell))\\
		&\ \ \ \ +p_0(\theta_k)(\theta_1-\theta_m-(\theta_2-\theta_m))>0
	\end{align*}
	so clearly $\eta$ is suboptimal and there exists an optimal solution $\eta^\star$ with $\eta_u^\star=0$. Furthermore, if $k>1$, $\ell>1$ and $k \neq l$, highly similar arguments can be used to show that $\eta_{(k,l)}^\star=\eta_{(k,l,k)}^\star=\eta_{(1,k,l)}^\star=\eta_{(k,1,\ell)}^\star=\eta_{(k,\ell,1)}^\star=0$. It can be concluded that for any $k>1$, $l>1$, $k\neq l$ if both $k\in u$ and $\l \in u$ then $\eta_u^\star=0$. With this in mind, along with the pruned constraints, solving the optimization problem \eqref{eq:noisyraw} comes down to, for all $k \notin u^\star$, identifying for which $u$ such that $k \in u$ we have $\eta_u>0$. Proving the general statement of the solution in equation \eqref{eq:noisy_solution} then comes down to showing that if $k\in u$ $\eta_{u}^\star=0$ unless $u\in \mathcal{U}_1(k)$. In other words, we wish to show that $\eta_{(k,k,k)}^\star=0$ (which is trivial, as $\mu((k,k,k))<\mu((k))$ but the two policies carry the same information about arm $k$) and that $u_1=1 \implies \eta_u^\star=0$. When $u^\star(\theta)=(1,2)$, we also need to show $\eta_{(k)}=0$ but this is completely analogous to Theorem \ref{the:CMOsolution}. 
	
	First, analogous to $u=(k,k,k)$, we easily find $\eta_{(1,k,k)}^\star=0$. Next, consider $u=(1,k,1)$. We find that redistributing the weight to $(1,k)$ gives less regret per average observation of arm $k$, as
	\begin{align*}
		&\ \frac{\mu^\star-\mu(1,1,k)}{p_0(\theta_1)I(\theta_k,\bar{\theta})}-\frac{\mu^\star-\mu(1,k,1)}{(1-p_0(\theta_1))I(\theta_k,\bar{\theta})}\\
		&= -\frac{(1-2\varepsilon)(\theta_1(1-\theta_2)+\theta_2(1-\theta_1))}{p_0(\theta_1)(1-p_0(\theta_1))I(\theta_k,\bar{\theta})}<0
	\end{align*}
	and so $\eta_u^\star=0$. Finally, consider $u=(1,k)$. For any $\eta$ with $\eta_u>0$, consider $\eta'$ with $\eta'_u = 0$, $\eta'_v=\eta_v + \frac{p_0(\theta_1)I(\theta_k,\bar{\theta})}{KL(\nu_{\theta}(u)||\nu_{(\theta^{(-k)},\bar{\theta})}(u))}\eta_u$ for $v=(1,k)$ and $\eta_w'=\eta_w$ for all $w\neq u,v$. Clearly, $\eta'$ is feasible and the difference in objective functions is $\eta_u$ multiplied by
	\begin{align*}
	&(\mu^*-\mu((1,k)))-\frac{p_0(\theta_1)I(\theta_k,\bar{\theta})}{KL(\nu_{\theta}(u)||\nu_{(\theta^{(-k)},\bar{\theta})}(u))}(\mu^*-\mu((k,1))) \\
	&= (\mu^*-\mu((1,k)))\left(1-\frac{p_0(\theta_1)I(\theta_k,\bar{\theta})}{KL(\nu_{\theta}(u)||\nu_{(\theta^{(-k)},\bar{\theta})}(u))}\right).
	\end{align*}
Note that 
\begin{align*}
&KL(\nu_{\theta}(u)||\nu_{(\theta^{(-k)},\bar{\theta})}(u))=(1-\varepsilon)\theta\log \left(\frac{\theta_k}{\bar{\theta}}\right) \\
&\ \ \ \  + \varepsilon(1-\theta_k)\log\left(\frac{1-\theta_k}{1-\bar{\theta}} \right)+p_0(\theta_k)\log\left(\frac{p_0(\theta_k)}{p_0(\bar{\theta})}\right).
\end{align*}
It thus suffices to prove that for $\theta_k<\theta_2$, $f(\theta_k):=p_0(\theta_1)I(\theta_k,\bar{\theta})KL(\nu_{\theta}(u)||\nu_{(\theta^{(-k)},\bar{\theta})}(u))\leq 0$. Since, by the properties of the KL-divergence, $f(\bar{\theta})=f'(\bar{\theta})=0$, this follows if for all $\theta_k<\bar{\theta}$, $f''(\bar{\theta})=0$. 

	We evaluate $f''(\theta_k)$ for $0<\theta_k<\bar{\theta}$. We have that $\frac{\partial^2}{\partial \theta_k^2}\theta_k\ln(\frac{\theta_k}{\theta_2}) = \frac{1}{\theta_k}$ and that  $\frac{\partial^2}{\partial \theta_k^2}(1-\theta_k)\ln(\frac{1-\theta_k}{1-\theta_2}) = \frac{1}{1-\theta_k}$. From this, it is relatively straightforward to find that $\frac{\partial^2}{\partial \theta_k^2}(\varepsilon\theta_k+(1-\varepsilon)(1-\theta_k))\ln(\frac{(\varepsilon\theta_k+(1-\varepsilon)(1-\theta_k))}{\varepsilon\theta_2+(1-\varepsilon)(1-\theta_2)}) = \frac{(1-2\varepsilon)^2}{(\varepsilon\theta_k+(1-\varepsilon)(1-\theta_k))}$. Then, we can write
	\begin{align*}
	f''(\theta_k)&=\frac{1}{\theta_k}(\varepsilon\theta_1+(1-\varepsilon)(1-\theta_1)-(1-\varepsilon))\\
	&+ \frac{1}{1-\theta_k} (\varepsilon\theta_1+(1-\varepsilon)(1-\theta_1)-\varepsilon)\\
	&- \frac{(1-2\varepsilon)^2}{(\varepsilon\theta_k+(1-\varepsilon)(1-\theta_k))}\\
	&\leq (1-2\varepsilon)\left(\frac{1-\theta_1}{1-\theta_k}-\frac{\theta_1}{\theta_k}\right)\leq 0
	\end{align*}
	where the first inequality is removal of a non-positive term and the second inequality comes from $\theta_k<\theta_1$ as well as $\varepsilon \leq \frac{1}{2}$. Since we have eliminated all possibilities, we have now found that $u_1=1 \implies \eta_u^\star=0$, which directly leads to the result in Theorem \ref{the:NCMO-solution}.
\end{proof}

	\section{Proof of Theorem \ref{the:SinglePredictiveUpperBound}}
	\label{appx:single_predictive_upper_bound}
	Our proof strategy is similar to that used in Combes et al. \cite{combes2015learning} or other analyses of the regret of bandit algorithms. Namely, we decompose the set of rounds into several subsets, and upper bound the regret generated in each of the subsets. In the following lemma, we show that thanks to the aggressive exploitation behavior of SPAM, the expected number of rounds where the leading policy is not $u^\star(\theta)$ is finite. 

\begin{lemma}\label{lem1}
		Choose $\delta \in (0,\delta_0)$, with $\delta_0$ defined in the statement of Theorem \ref{the:SinglePredictiveUpperBound}. We define the following sets:
		\begin{align*}
			\mathcal{A} &= \{t\in\mathbb{N}:\mathcal{L}(t)\neq u^\star(\theta)\}\\
			\mathcal{D} &= \{t\in\mathbb{N}:(\exists i \in \mathcal{L}(t): |\hat{\theta}_i(t)-\theta_i|\geq \delta)\}
		\end{align*}
		and $\mathcal{C}=\mathcal{A}\cup \mathcal{D}$. Furthermore, we denote $\beta = (1-\theta_1)^{-1}$ Then, under Algorithm \ref{alg:SPAM}, we have
		\begin{equation}
			\E{|\mathcal{C}|} \leq 4K[4(K+\beta^2)+\delta^{-2}(\beta+1)] + 30.
		\end{equation}
\end{lemma}

{\vskip 0.3cm}
\noindent
{\it Proof of Lemma \ref{lem1}.}	Introduce the sets
		\begin{align*}
			\mathcal{E} &= \{t \in \mathbb{N}: (\exists i\in u^\star(\theta): b_i(t)\leq \theta_i)\}\\
			\mathcal{G} &= \{t \in \mathcal{A}\backslash(\mathcal{D}\cup\mathcal{E}): (\exists i \in u^\star(\theta): i \notin \mathcal{L}(t),\\
			&\ \ \ \ \ \ \ \ \ \ \ \ \ \ \ \ \ \ \ \ \ \ \ \ \ \  |\hat{\theta}_i(t)-\theta_i|> \delta)\}.
		\end{align*}
		We will show that $\mathcal{C} \subseteq \mathcal{D} \cup \mathcal{E} \cup \mathcal{G}$. Take $t\in \mathcal{A}$ which does not fulfill $\forall i \in u^\star(\theta)\backslash \mathcal{L}(t): |\hat{\theta}_i(t)-\theta_i|\leq \delta$. Clearly, $t \in \mathcal{E} \cup \mathcal{G}$. Now take $t\in \mathcal{A}$ such that this is fulfilled. First, treat the case $u^\star(\theta)=(1,2)$. Since $t \in \mathcal{A}$, either (a) $\mathcal{L}(t)=(1)$ or (b) there exists $i,j$ such that $i \in u^\star(\theta)\backslash \mathcal{L}(t)$, $\theta_i>\theta_j$ and $\hat{\theta}_j(t)>\hat{\theta}_i(t)$. If (a) is true we have that $\frac{c}{1-\hat{\theta}_1(t)}>\hat{\theta}_2(t)\geq \theta_2-\delta$ where the second inequality follows from $2 \in u^\star(\theta)\backslash \mathcal{L}(t)$. But the definition of $\delta_0$ then implies that $\hat{\theta}_1(t)>\theta_1 + \delta$, so $t \in \mathcal{D}$. If (b) is true we have $\hat{\theta}_j(t)>\hat{\theta}_i(t)\geq \theta_i-\delta>\theta_j+\delta$ where the last inequality follows from the definition of $\delta_0$. Therefore we have $t \in \mathcal{D}$.
		
		 Next, treat the case $u^\star(\theta) = 1$. Then, since $t\in\mathcal{A}$ there exists $i \in \mathcal{L}(t)$ such that either (a) $\hat{\theta}_i(t) \geq \frac{c}{1-\hat{\theta}_1(t)}$, or (b) $\hat{\theta}_i(t)\geq \hat{\theta}_1(t)$. In both cases, either $t \in \mathcal{D}$ or $|\hat{\theta}_1(t)-\theta_1|\leq \delta$. We focus on the case $|\hat{\theta}_1(t)-\theta_1|\leq \delta$. If (a) is true, $\hat{\theta}_i(t)\geq \frac{c}{1-\hat{\theta}_1(t)}\geq \frac{c}{1-\theta_1+\delta}>\theta_i+\delta$ with the strict inequality following from the definition of $\delta_0$, thus we have $t \in \mathcal{D}$. If (b) is true, $t \in \mathcal{D}$ with the same reasoning as in the case $u^\star(\theta)=(1,2)$. No matter what, we have $t \in \mathcal{D}$, or in other words, $\mathcal{C} \subseteq \mathcal{D} \cup \mathcal{E} \cup \mathcal{G}$.
		
		Now, we wish to bound $\E{|\mathcal{D}|}$, $\E{|\mathcal{E}|}$ and $\E{|\mathcal{G}|}$. The result will follow by a union bound.
		
		Decompose $\mathcal{D}=\bigcup_{i=1}^K \mathcal{D}_i$, where $\mathcal{D}_i = \{t \in \mathbb{N}: i\in \mathcal{L}(t), |\hat{\theta}_i(t)-\theta_i|\geq\delta\}$. Note that by the definition of the algorithm, the probability of observing arm $i$ given that $i \in \mathcal{L}(n)$ (and therefore, given that $t\in \mathcal{D}_i$), is at least $\frac{\beta^{-1}}{2}$. Thus, by Lemma 5 of Combes et al. \cite{combes2015learning} with $H=\mathcal{D}_1$ and $c=\frac{\beta^{-1}}{2}$ we have that $\E{|\mathcal{D}_i|} \leq 4\beta[4\beta+\delta^{-2}]$ and by a union bound
		\begin{equation*}
			\E{|\mathcal{D}|} \leq 4K\beta[4\beta+\delta^{-2}].
		\end{equation*}

		Next, for any $i \in u^\star(\theta)$, let $\mathcal{E}_i = \{ t \in \mathbb{N}: b_i(t) \leq \theta_i\}$. It follows that $\mathcal{E} = \bigcup_{i\in u^\star(\theta)}\mathcal{E}_i$. By Lemma 6 of Combes et al., we have that $\E{|\mathcal{E}_i|}\leq 15$ and thereby, by a union bound,
		\begin{equation*}
		\E{|\mathcal{E}|} \leq 2*15=30,
		\end{equation*}
		since there can be at most 2 distinct elements in $u^\star(\theta)$.
		
		Finally, for any $i \in u^\star(\theta)$, let $\mathcal{G}_i = \{t \in \mathcal{A}\backslash(\mathcal{D}\cup\mathcal{E}): i \notin \mathcal{L}(t), |\hat{\theta}_i(t)-\theta_i| >\delta\}$. Then $\mathcal{G} = \bigcup_{i\in u^\star(\theta)}\mathcal{G}_i$.
		
		Consider $i=1$ and choose $t \in \mathcal{G}_1$. Then, since $t \notin \mathcal{E}$ we have $b_1(t)\geq \theta_1$. Furthermore, since $t \notin \mathcal{D}$ there exists $j \in \mathcal{L}(t):j>1$ such that
		\begin{equation*}
			\hat{\bar{\theta}}(t)\leq\hat{\theta}_j(t)\leq \theta_j+\delta \leq \frac{\theta_j+\theta_1}{2}<\theta_1\leq b_1(t).
		\end{equation*}
		Thus, we have $1\in\mathcal{B}(t)$.
		
		Consider now $i=2$ (in which case $u^\star(\theta) = (1,2)$) and choose $t \in \mathcal{G}_2$. Then, since $t\in \mathcal{A}$, either there exists $j>2$ such that $j \in \mathcal{L}(t)$ (in which case, the exact same argument as in the case $i=1$ applies) or $\mathcal{L}(t)=(1)$. In the latter case, since $t \notin \mathcal{E}$ we have $b_2(t)\geq \theta_2$ and since $t \notin \mathcal{D}$ we have $|\hat{\theta}_1(t)-\theta_1|\leq \delta$. Thus, by definition of $\delta_0$
		\begin{equation*}
			\hat{\bar{\theta}}(t) = \frac{c}{1-\hat{\theta}_1(t)}\leq \frac{c}{1-\theta_1-\delta}<\theta_2\leq b_2(t).
		\end{equation*}
		Thus, no matter what, $i \in \mathcal{B}(t)$. By the definition of the algorithm, the probability of observing $i$ given that $i \in \mathcal{B}(t)$ is at least $\frac{1}{2K}$. Then, we can once again employ Lemma 5 of Combes et al. with $H=\mathcal{G}_i$ and $c=\frac{1}{2K}$ to find
		that $\E{|\mathcal{G}_i|} \leq 4K(4K+\delta^{-2})$. This immediately yields
		\begin{equation*}
			\E{|\mathcal{G}|}\leq 4K[4K+\delta^{-2}].
		\end{equation*}
		
		By a union bound, we find
		\begin{align*}
			\E{|C}| \leq &\E{|\mathcal{D}|} + \E{|\mathcal{E}|} + \E{|\mathcal{G}|}\\ &\leq 4K[4(K+\beta^2)+\delta^{-2}(\beta+1)] + 30,
		\end{align*}
		which is the desired result.\ep
{\vskip 0.3cm}
\noindent
{\it Proof of Theorem \ref{the:SinglePredictiveUpperBound}.} Define $\mathcal{K}_i^1=\{t\in [1,T]: t \notin \mathcal{C}, u(t)=(i,1)\}$ and $\mathcal{K}_i^2=\{t\in [1,T]: t \notin \mathcal{C}, u(t)=(i)\}$. By design of the algorithm, if $t \notin \mathcal{C}$, the algorithm will either play the optimal policy or it will play $(i,1)$ or $(i)$ for some $i \notin u^\star(\theta)$. Since $\mu^\star-\mu(u)\leq 1+c$ for all $u$, we can decompose the regret as
		\begin{align*}
			R^{\mathrm{SPAM}}_\theta(T) \leq (1+c)\E{|\mathcal{C}|} &+ \sum_{i\notin u^\star(\theta)} [\mu^\star-\mu((i,1))]\E{|\mathcal{K}_i^1|}\\
			 &+ \sum_{i\notin u^\star(\theta)} [\mu^\star-\mu((i))]\E{|\mathcal{K}_i^2|}.
		\end{align*}
		We now bound $\E{|\mathcal{K}_i^1|}$ (the bound on $\E{|\mathcal{K}_i^2|}$ will be analogous). Recall that $g(\theta_1,\theta_2,\delta)$ is defined such that
		\begin{equation*}
			g(\theta_1,\theta_2,\delta) := \begin{cases}
			\theta_1 - \delta, &\bar{\theta}=\theta_{1}\\
			\theta_2 - \delta, &\bar{\theta}=\theta_{2}\\
			\frac{c}{1-\theta_{1}+\delta}, &\bar{\theta}=\frac{c}{1-\theta_1}.
			\end{cases}
		\end{equation*}
		Note that $g(\theta_{1},\theta_{2},0) = \bar{\theta}$. Now, choose $\epsilon \in (0,1)$, define the number of elements in $\mathcal{K}_i^1$ up to time $t$ as $k_i(t):=\sum_{s=1}^{t}\mathds{1}(s\in\mathcal{K}_i^1)$, and define $n_0=\frac{f(T)}{I(\theta_i+\delta,g(\theta_{1},\theta_{2},\delta))}$. Then, we wish to decompose $\mathcal{K}_i^1$ into $\mathcal{K}_{i,1}^1\cup\mathcal{K}_{i,2}^1$, where
		\begin{align*}
			\mathcal{K}_{i,1}^1 &= \{t\in \mathcal{K}_i^1: n_i(t)\leq (1-\epsilon)k_i(t)\ \text{or}\ |\hat{\theta}_i(t)-\theta_i|\geq\delta\}\\
			\mathcal{K}_{i,2}^1 &= \{t\in \mathcal{K}_i^1: n_0\geq (1-\epsilon)k_i(t)\ \text{and}\ |\hat{\theta}_i(t)-\theta_i|<\delta\}.
		\end{align*}
		Now we show that this decomposition is valid, by contradiction. Take $t$ in $\mathcal{K}_i^1\backslash(\mathcal{K}_{i,1}^1\cup \mathcal{K}_{i,2}^1)$. Since $t \notin \mathcal{K}_{i,1}^1$, $n_i(t) \geq (1-\epsilon)k_i(t)$ and since $t \notin \mathcal{K}_{i,2}^1$, $(1-\epsilon)k_i(t)\geq n_0$, so $n_i(t)\geq n_0$, which we call inequality (a). 
		
		Furthermore, since $t\in \mathcal{K}_i^1$ and by design of the algorithm, we get $i\in \mathcal{B}(t)$ which in turn implies $b_i(t)\geq \hat{\bar{\theta}}(t)$. Since $t \notin \mathcal{C}$ and $\delta < \delta_0$ we must have (by definition of $\delta_0$) $\hat{\bar{\theta}}(t) = g(\hat{\theta}_1(t),\hat{\theta}_2(t),0)\geq g(\theta_1,\theta_2,\delta)$. Therefore we have $b_i(t)\geq g(\theta_1,\theta_2,\delta)$, which we call inequality (b).
		
		Putting inequalities (a) and (b) together with the definition of $b_i(t)$ we obtain
		\begin{align*}
			n_0 I(\hat{\theta}_i(t),g(\theta_{1},\theta_{2},\delta)) &\leq n_i(t) I(\hat{\theta}_i,g(\theta_{1},\theta_{2},\delta)) \\
			&\leq f(t)\leq f(T)
		\end{align*}
		and so, by definition of $n_0$, we obtain $I(\hat{\theta}_i(t),g(\theta_{1},\theta_{2},\delta)) \leq I(\theta_i+\delta,g(\theta_{1},\theta_{2},\delta))$ which by monotonicity of $I(x,g(\theta_{1},\theta_{2},\delta))$ on the interval $[0,g(\theta_{1},\theta_{2},\delta)]$ implies that $\hat{\theta}_i(t) \geq \theta_i + \delta$. But then $t\in \mathcal{K}_{i,1}^1$, which is a contradiction. Therefore $\mathcal{K}_i^1\subseteq \mathcal{K}_{i,1}^1\cup \mathcal{K}_{i,2}^1$.
		
		Next, we bound $\E{|\mathcal{K}_{i,1}^1|}$ and $\E{|\mathcal{K}_{i,2}^1|}$. First, note that the probability of observing arm $k$ given that $t\in\mathcal{K}_{i}^1$ (or for that matter, given that $t\in \mathcal{K}_i^2$) is $1$. Next, we can use Corollary 1 in \cite{combes2015learning} with $H=\mathcal{K}_{i,1}^1$ and $c=1$ to bound $\E{|\mathcal{K}_{i,1}^1|}\leq \epsilon^{-2}+(1-\epsilon)^{-1}\delta^{-2}$.
		
		Finally, note that by definition of $\delta_0$, if it is true that $(1-\theta_i)\theta_1<c$ then, if $t \in \mathcal{K}_{i,2}^1$ we have $(1-\hat{\theta}_1(t))\hat{\theta}_i(t)\leq (1-\theta_1+\delta)(\theta_i+\delta)<c$ since $\theta_i < \theta_2$. But by design of the algorithm, $u(t) \neq (k,1)$ which is a contradiction. Therefore, if $(1-\theta_i)\theta_1<c$, it follows that $\E{|\mathcal{K}_{i,2}^1|}=0$. We also have that if $t \in \mathcal{K}_{i,2}^1$ then $k_i(t)\leq (1-\epsilon)^{-1}n_0$. Since $k_i(t)$ is incremented at $t$, we then have that $\E{|\mathcal{K}_{i,2}^1|}\leq (1-\epsilon)^{-1}n_0$ and in total
		\begin{equation*}
			\E{|\mathcal{K}_{i,2}^1|}\leq \mathds{1}((1-\theta_i)\theta_1\geq c)(1-\epsilon)^{-1}n_0
		\end{equation*}

		Now we put everything together (with an analogous bound on $\E{|\mathcal{K}_i^2|}$) and we obtain
		\begin{align*}
			R^{SPAM}_\theta(T) &\leq (1+c)(4K[4(K+\beta^2)+\delta^{-2}(\beta+1)] + 30)\\
			&+2(1+c)K[\epsilon^{-2}+(1-\epsilon)^{-1}\delta^{-2}]\\
			&+\sum_{i\notin u^\star(\theta)}\frac{\mathds{1}((1-\theta_i)\theta_1\geq c)(\mu^\star-\mu((i,1)))}{(1-\epsilon)I(\theta_i,g(\theta_1,\theta_2,\delta))}f(T)\\
			&+\sum_{i\notin u^\star(\theta)} \frac{\mathds{1}((1-\theta_i)\theta_1< c)(\mu^\star-\mu((i)))}{(1-\epsilon)I(\theta_i,g(\theta_1,\theta_2,\delta))}f(T)\\
			&= 2K(1+c)[8(K+\beta^2)+\epsilon^{-2}\\
			&\ \ \ \ +\delta^{-2}(2(\beta+1)+(1-\epsilon)^{-1})]\\
			&\ \ \ \ + \sum_{i\notin u^\star(\theta)} \frac{H_i(\theta)I(\theta_i,\bar{\theta})}{(1-\epsilon)I(\theta_i,g(\theta_1,\theta_2,\delta))}f(T)
		\end{align*}
		which directly leads to the result of Theorem \ref{the:SinglePredictiveUpperBound}.
		
\ep

	\bibliography{ref}{}
\bibliographystyle{IEEEtran}

\end{document}